\documentclass{elsarticle}

\usepackage{lineno,hyperref}
\usepackage{comment}
\usepackage{fancyvrb}
\usepackage{graphicx}
\usepackage{hyperref}
\usepackage[ragged]{footmisc}
\usepackage{ragged2e}
\usepackage[section]{placeins}
\usepackage{layout}
\usepackage{floatpag}
\usepackage{gensymb}
\usepackage{lscape}
\usepackage{geometry}
\usepackage[T1]{fontenc}
\usepackage{setspace}

\modulolinenumbers[5]

\begin{document}

\title{Mathematics as information compression via the matching and unification of patterns}

\author{J Gerard Wolff\footnote{Dr Gerry Wolff, BA (Cantab), PhD (Wales), CEng, MBCS, MIEEE; CognitionResearch.org, Menai Bridge, UK; \href{mailto:jgw@cognitionresearch.org}{jgw@cognitionresearch.org}; +44 (0) 1248 712962; +44 (0) 7746 290775; {\em Skype}: gerry.wolff; {\em Web}: \href{http://www.cognitionresearch.org}{www.cognitionresearch.org}.}}

\maketitle

\begin{abstract}

\sloppy This paper describes a novel perspective on the foundations of mathematics: how mathematics may be seen to be largely about `information compression via the matching and unification of patterns' (ICMUP). ICMUP is itself a novel approach to information compression, couched in terms of non-mathematical primitives, as is necessary in any investigation of the foundations of mathematics. This new perspective on the foundations of mathematics reflects the idea that, as an aid to human thinking, mathematics should conform to much evidence for the importance of information compression in human learning, perception, and cognition. This new thinking about the nature of mathematics has grown out of an extensive programme of research developing the {\em SP Theory of Intelligence} and its realisation in the {\em SP Computer Model}, a system in which a generalised version of ICMUP---the powerful concept of SP-multiple-alignment---plays a central role. These ideas may be seen to be part of a {\em Big Picture} comprising six areas of interest, with information compression as a unifying theme. The paper describes the close relation between mathematics and information compression, and describes examples showing how variants of ICMUP may be seen in widely-used structures and operations in mathematics. Examples are also given to show how the mathematics-related disciplines of logic and computing may be understood as ICMUP. Also discussed is how the intimate relation between information compression and concepts of inference and probability may be seen as a driver for the making of mathematical inferences, how it relates to the already-established view that, at a fundamental level, mathematics is intrinsically probabilistic, and how that latter view may be reconciled with the all-or-nothing, `exact', forms of calculation or inference that are familiar in mathematics. There are many potential benefits and applications of these ideas.

\end{abstract}


\maketitle

\section{Introduction}\label{introduction_section}

This paper, which draws on and considerably expands some of the thinking in \cite[Chapter 10]{wolff_2006}, describes how much of mathematics, perhaps all of it, may be seen as a set of techniques for the compression of information, and their application.

For reasons given in Section \ref{icmup_section}, information compression is seen here as a process of searching for patterns that match each other and merging or `unifying' patterns that are the same.\footnote{Here and throughout this paper, the term `unification' will mean a simple merging of two or more matching patterns to make one. This use of the term is only loosely related to the meaning of the term `unification' in logic and computer science.} The expression `information compression via the matching and unification of patterns' is abbreviated as `ICMUP'. The main subject of this paper---mathematics as ICMUP---is referred to as `MICMUP'.

This MICMUP perspective appears to be novel. It has apparently not been previously described in writings about the philosophy of mathematics, the philosophy of science, or elsewhere.

The MICMUP perspective has grown out of a programme of research developing the {\em SP Theory of Intelligence} and its realisation in the {\em SP Computer Model} \cite{sp_extended_overview,wolff_2006}, seeking to simplify and integrate observations and concepts across artificial intelligence, mainstream computing, mathematics, and human learning, perception, and cognition. The SP system is outlined in Section \ref{sp_theory_section} and described in a little more detail in \ref{sp_system_appendix}, with pointers to where fuller information may be found.

MICMUP and the SP system may be seen to be part of a `Big Picture' with information compression as a unifying theme, described in outline in Section \ref{the_big_picture_section}.

\subsection{Presentation}\label{presentation_section}

The next section provides some background to the main body of the paper including: the novelty of `mathematics as information compression'; mathematics as an aid to human thinking may be seen to reflect the importance of information compression in human learning, perception, and cognition; a sketch of the {\em SP Theory of Intelligence} and its realisation in the {\em SP Computer Model}; an outline of a `Big Picture' of which this research is a part; ICMUP as an alternative approach to information compression; and an outline of seven variants of ICMUP.

Section \ref{mathematics_as_icmup_section} describes, first, how information compression may be seen in the workings of mathematics, and then how variants of ICMUP may be seen in the structure and workings of mathematics.

Section \ref{mathematics-related_disciplines_as_icmup_section} outlines how the mathematics-related disciplines of  logic and computing may be seen as ICMUP.

Section \ref{ic_inference_probabilities_section} discusses briefly how MICMUP is in keeping with other evidence that mathematics is fundamentally probabilistic, and how this may be reconciled with the all-or-nothing, `exact', forms of calculation or inference that are familiar in mathematics.

Section \ref{so_what_section} outlines some potential benefits and applications of MICMUP, and of ideas that are associated with MICMUP in the Big Picture.

\section{Background}

This section outlines some preliminaries to the sections that follow.

\subsection{The novelty of the idea that mathematics may be seen as compression of information}\label{novelty_of_maths_as_ic_section}

Three recent books about the philosophy of mathematics \cite{linnebo_2017,marcus_mcevoy_2016,parsons_2014} make no mention of anything resembling information compression or ICMUP. More generally, the idea that information compression might be part of the foundations of mathematics appears to be invisible in writings about the nature of mathematics.

Keith Devlin's academic book, {\em Logic and Information} \cite{devlin_1991}, aims to develop a mathematical theory of information, a goal which is related to but distinct from the central idea in MICMUP, that mathematics may be seen to be largely ICMUP.

Devlin's later book for the general reader, {\em Mathematics: The Science of Patterns} \cite{devlin_1997}, discusses things like ``patterns of symmetry [such as] the symmetry of a snowflake or a flower'' (p.~145) and ``the patterns involved in packing objects in an efficient manner'' (p.~152) which would be relatively complex kinds of pattern representing abstract concepts. But the key ideas in MICMUP, as described in this paper, are not discussed.

\sloppy Amongst the several ``isms'' in the philosophy of mathematics---foundationism, logicism, intuitionism, formalism, Platonism, neo-Fregeanism, and more---the three which are perhaps most closely related to MICMUP are: {\em psychologism} (mathematical concepts derive from human psychology); {\em embodied mind theories} (mathematical thought is a natural outgrowth of human cognition); and {\em intuitionism} (mathematics is a creation of the human mind). This is because the latter three views are broadly consistent with the afore-mentioned evidence that much of HLPC may be understood as information compression \cite{sp_compression}. But it appears that there is nothing like information compression or ICMUP in any of those three views or any other school of thought in the philosophy of mathematics.

\subsection{Mathematics as an aid to human thinking, and evidence for Information compression in human cognition}\label{ic_in_hlpc_section}

Since mathematics may be seen as an aid to human thinking, it should not be surprising to find that it conforms to much evidence for the importance of information compression, and more specifically ICMUP, in human learning, perception, and cognition \cite{sp_compression}.

\subsection{Outline of the SP Theory of Intelligence and the SP Computer Model}\label{sp_theory_section}

The ideas and arguments presented in this paper have grown out of an extensive programme of research developing the {\em SP Theory of Intelligence} and its realisation in the {\em SP Computer Model}. This subsection highlights some key ideas in this research. There is more detail about the SP system in \ref{sp_system_appendix} with pointers to where fuller information may be found.

The main aim in this research is, in accordance with Ockham's razor: simplification and integration of observations and concepts across artificial intelligence, mainstream computing, mathematics, and human leaning, perception, and cognition (HLPC), with ICMUP as a unifying theme.\footnote{Since people often ask what the name ``SP'' stands for, it is short for {\em Simplicity} and {\em Power}, two ideas that, together, mean the same as information compression. This is because compression of a body of information, {\bf I}, means maximising the {\em simplicity} of {\bf I} by reducing, as much as possible, repetition of information or {\em redundancy} in {\bf I}, whilst retaining a percentage of its non-redundant descriptive or explanatory {\em power}: 100\% in the case of lossless compression, or some smaller percentage in the case of lossy compression.}

The focus on information compression and ICMUP is because there is substantial evidence that much of HLPC may be understood in those terms \cite{sp_compression}.

An important idea in the SP system is the powerful concept of {\em SP-multiple-alignment}, borrowed and adapted from the concept of `multiple sequence alignment' in bioinformatics, and outlined in \ref{sp-multiple-alignment_appendix}. The SP-multiple-alignment concept, which may be seen as a generalised version of ICMUP, is the key to the SP system's versatility in modelling diverse aspects of human intelligence, in the representation of diverse kinds of knowledge, and in the seamless integration of diverse aspects of intelligence and diverse forms of knowledge, in any combination.

Some related issues are described briefly in \ref{related_issues_appendix} with pointers to where fuller information may be found.

\subsection{The `Big Picture'}\label{the_big_picture_section}

This research may be seen to be part of a `Big Picture' with (at least) six components:

\begin{itemize}

    \item {\em Information compression as a foundation for mathematics}. The present paper, ``Mathematics as Information Compression Via the Matching and Unification of Patterns,'' argues that much of mathematics, perhaps all of it, may be understood in terms of ICMUP.

    \item {\em Information compression as a unifying principle in science}. It is widely agreed that ``Science is, at root, just the search for compression in the world'' \cite[p.~247]{barrow_1992}, with variations such as ``Science may be regarded as the art of data compression'' \cite[p.~585]{li_vitanyi_2014}.

    \item {\em Information compression and concepts of inference and probability}. It is known that there is an intimate relation between information compression and concepts of inference and probability (Section \ref{ic_inference_probabilities_section}).

    \item {\em Evidence for information compression as a unifying principle in HLPC}. A companion to the present paper describes relatively direct empirical evidence for information compression, which in many cases may be seen as ICMUP, as a unifying principle in HLPC \cite{sp_compression}. In view of that evidence, and since mathematics has been developed as an aid to human thinking, it should not be surprising that mathematics may be founded on information compression (Section \ref{ic_maths_hlpc_section}).

    \item {\em Information compression in the SP Theory of Intelligence}. A central idea in the SP Theory of Intelligence (Section \ref{sp_theory_section} and \ref{sp_system_appendix}) is the powerful concept of {\em SP-multiple-alignment} which may be seen to be a generalised version of ICMUP (Section \ref{sp-multiple-alignment_section} and \cite[Appendix B]{sp_software_engineering}).

    \item {\em Information compression in neuroscience}. Because of its central role in the SP system, ICMUP is likely to prove significant in {\em SP-Neural} \cite{spneural_2016}, a version of the SP theory which describes how abstract concepts in the theory may be realised in terms of neurons and their interconnections.

\end{itemize}

The six components of the Big Picture are mutually supportive in the sense that the credibility of any one of them, including the main thesis of this paper, is strengthened via empirical and analytical evidence in support of the Big Picture across all six of its components.

The significance of the Big Picture is discussed briefly in Section \ref{so_what_section}.

\subsection{Information compression via the matching and unification of patterns}\label{icmup_section}

Readers who are acquainted with techniques for the compression of information will know that many of them, such as Huffman coding, arithmetic coding, and wavelet compression, have a mathematical flavour (see, for example, \cite{sayood_2012}). Much the same may be said about algorithmic compression in the framework of AIT \cite{li_vitanyi_2014}.

Since ideas of that kind have a good pedigree and have proved their worth in many applications, one might suppose that they would be the starting point for any discussion of how mathematics may be understood in terms of information compression. But:

\begin{itemize}

    \item The SP programme of research attempts to reach down below the mathematics of other approaches, to focus on the relatively simple, `primitive' idea that information compression may be understood in terms of the matching and unification of patterns.

    \item In any discussion of the fundamentals of mathematics, it would not be appropriate to use mathematics itself.

    \item Since ICMUP is a relatively `concrete' idea, less abstract than much of mathematics, it suggests avenues that may be explored in understanding possible mechanisms for information compression in artificial systems and in brains and nervous systems.

\end{itemize}

ICMUP and its variants may appear too childishly simple to merit attention in any discussion of the fundamentals of mathematics. But ICMUP is bedrock in the powerful concept of SP-multiple-alignment, outlined in \ref{sp-multiple-alignment_appendix}, which has proven capabilities, not only in modelling the six other versions of ICMUP described in Section \ref{seven_techniques_for_icmup_section} (as described in \cite[Appendix B]{sp_software_engineering}), but also, more importantly, in modelling diverse aspects of human intelligence, the representation of diverse forms of knowledge, and the seamless integration of diverse aspects of intelligence and diverse forms of knowledge, in any combination \cite{sp_extended_overview,wolff_2006}. And, as described in \cite{sp_compression}, ICMUP is prominent in HLPC.

\subsection{Seven techniques for the compression of information via the matching and unification of patterns}\label{seven_techniques_for_icmup_section}

While care has been taken in this programme of research to avoid unnecessary duplication of information across different publications, the importance of the following seven variants of ICMUP has made it necessary, for the sake of clarity, to describe them quite fully both in this paper and also in \cite{sp_compression}.

\subsubsection{Basic ICMUP: information compression via the matching and unification of patterns}\label{basic_icmup_section}

The simplest of the techniques to be described is to find two or more patterns that match each other within a body of information, {\bf I}, and then merge or `unify' them so that multiple instances are reduced to one. This is illustrated in the upper part of Figure \ref{unification_figure} where two instances of the pattern `\texttt{INFORMATION}' near the top of the figure has been reduced to one instance, shown in the middle of the figure, with `\texttt{w62}' appended at the front, for reasons given in Section \ref{chunking-with-codes_section}, below.

Here, and in subsections below, we shall assume that the single pattern which is the product of unification is placed in some kind of dictionary of patterns that is separate from {\bf I}.

The version of ICMUP just described will be referred to as {\em basic ICMUP}.

A detail that should not distract us from the main idea is that, when compression of a body of information, {\bf I}, is to be achieved via basic ICMUP, any repeating pattern that is to be unified should occur more often in {\bf I} than one would expect by chance.

\subsubsection{Chunking-with-codes}\label{chunking-with-codes_section}

A point that has been glossed over in describing basic ICMUP is that, when a body of information, {\bf I}, is to be compressed by unifying two or more instances of a pattern like `\texttt{INFORMATION}', there is a loss of information about the location within {\bf I} of each instance of `\texttt{INFORMATION}'. In other words, basic ICMUP achieves `lossy' compression of {\bf I}.

\begin{figure}[H]
\centering
\includegraphics[width=0.8\textwidth]{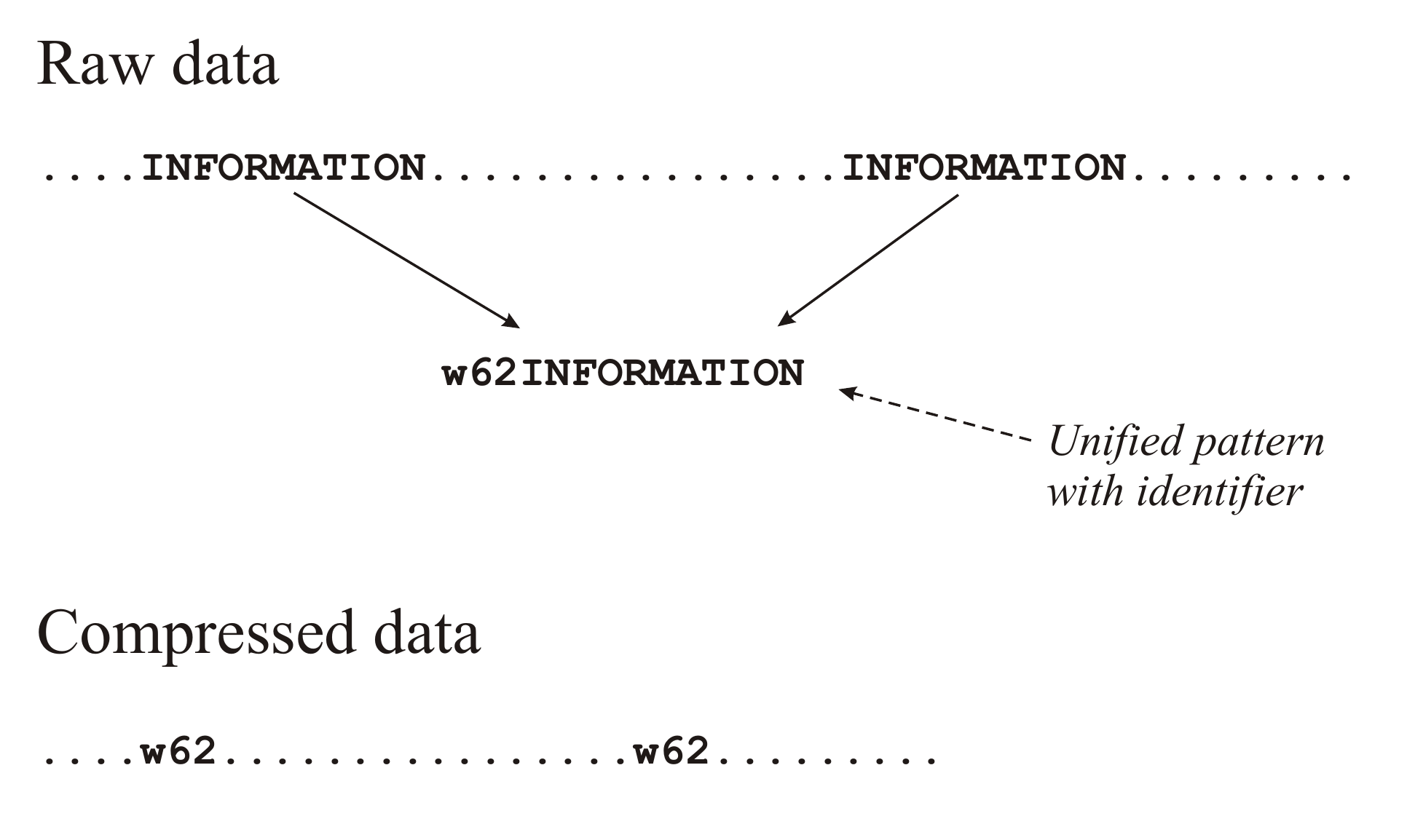}
\caption{A schematic representation of the way two instances of the pattern `\texttt{INFORMATION}' in a body of data may be unified to form a single `unified pattern', with `w62' as an identifier assigned by the system. The lower part of the figure shows how the data may be compressed by replacing each instance of `\texttt{INFORMATION}' with a copy of the corresponding identifer. Reproduced with permission from Figure 2.3 in \protect\cite{wolff_2006}.}
\label{unification_figure}
\end{figure}

This problem is overcome in the {\em chunking-with-codes} variant of ICMUP:

\begin{itemize}

    \item A unified pattern like `\texttt{INFORMATION}', which is often referred to as a `chunk' of information,\footnote{There is a little more detail about the concept of `chunk' in \cite[Section 2.4.2]{sp_compression}.} is stored in a dictionary of patterns, as mentioned in Section \ref{basic_icmup_section}.

    \item As before, the unified chunk is given a relatively short name, identifier, or `code', like the `\texttt{w62}' pattern appended at the front of the `\texttt{INFORMATION}' pattern in the middle of Figure \ref{unification_figure}.

    \item Then the `\texttt{w62}' code is used as a shorthand which replaces the `\texttt{INFORMATION}' chunk of information wherever it occurs within {\bf I}. This is shown at the bottom of Figure \ref{unification_figure}.

    \item Since the code `\texttt{w62}' is shorter than each instance of the pattern `\texttt{INFORMATION}' which it replaces, the overall effect is to shorten {\bf I}. But, unlike basic ICMUP, chunking-with-codes achieves `lossless' compression of {\bf I}.

    \item A detail here is that compression can be optimised by giving shorter codes to chunks that occur frequently and longer codes to chunks that are rare. This may be done using some such scheme as Shannon-Fano-Elias coding, described in, for example, \cite{cover_thomas_1991}.

\end{itemize}

\subsubsection{Schema-plus-correction}\label{schema-plus-correction_section}

A variant of the chunking-with-codes version of ICMUP is called {\em schema-plus-correction}. Here, the `schema' is like a chunk of information and, as with chunking-with-codes, there is a relatively short identifier or code that may be used to represent the chunk.

What is different about the schema-plus-correction idea is that the schema may be modified or `corrected' in various ways on different occasions.

For example, a menu for a meal in a cafe or restaurant may be something like `\texttt{MN: ST MC PG}', where `\texttt{MN}' is the identifier or code for the menu, `\texttt{ST}' is a variable that may take values representing different kinds of `starter', `\texttt{MC}' is a variable that may take values representing different kinds of `main course', and `\texttt{PG}' is a variable that may take values representing different kinds of `pudding'.

With this scheme, a particular meal may be represented economically as something like `\texttt{MN: ST(st2) MC(mc5) PG(pg3)}', where `\texttt{st2}' is the code or identifier for `minestrone soup', `\texttt{mc5}' is the code for `vegetable lassagne', and `\texttt{pg3}' is the code for `ice cream'; another meal may be represented economically as `\texttt{MN: ST(st6) MC(mc1) PG(pg4)}', where `\texttt{st6}' is the code or identifier for `prawn cocktail', `\texttt{mc1}' is the code for `lamb shank', and `\texttt{pg4}' is the code for `apple crumble'; and so on. Here, the codes for different dishes serve as modifiers or `corrections' to the categories `\texttt{ST}', `\texttt{MC}', and `\texttt{PG}' within the schema `\texttt{MN: ST MC PG}'.

\subsubsection{Run-length coding}\label{run-length_coding_section}

A third variant, {\em run-length coding}, may be used where there is a sequence of two or more copies of a pattern, each one except the first following immediately after its predecessor like this:

\begin{quote}

    `\texttt{INFORMATIONINFORMATIONINFORMATIONINFORMATIONINFORMATION}'.

\end{quote}

In this case, the multiple copies may be reduced to one, as before, with something to say how many copies there are (eg, `\texttt{(INFORMATION)$\times5$}'), or where the repetition begins and ends (eg, `\texttt{[(INFORMATION)*]}' where `\texttt{[}' and `\texttt{]}' are the beginning and end symbols, and `\texttt{*}' signifies repetition), or, more vaguely, that the pattern is repeated without anything to say when the sequence stops (eg, `\texttt{(INFORMATION)*}').

In a similar way, a sports coach might specify exercises as something like ``touch toes ($\times 15$), push-ups ($\times 10$), skipping ($\times 30$), ...'' or ``Start running on the spot when I say `start' and keep going until I say `stop'\thinspace''.

With the `running' example, ``start'' marks the beginning of the sequence, ``keep going'' in the context of ``running'' means ``keep repeating the process of putting one foot in front of the other, in the manner of running'', and ``stop'' marks the end of the repeating process. It is clearly much more econonomical to say ``keep going'' than to constantly repeat the instruction to put one foot in front of the other.

\subsubsection{Class-inclusion hierarchies}\label{class-inclusion_hierarchies_section}

A widely-used idea in everyday thinking and elsewhere is the {\em class-inclusion hierarchy}: the grouping of entities into classes and the grouping of classes into higher-level classes through as many levels as are needed.

This idea may achieve ICMUP because, at each level in the hierarchy, attributes may be recorded which apply to that level and all levels below it. This can mean great economies because, for example, it is not necessary to record that cats have fur, dogs have fur, rabbits have fur, and so on---it is only necessary to record that mammals have fur and ensure that all lower-level classes and entities can `inherit' that attribute. In effect, multiple instances of the attribute `fur' have been merged or unified to create that attribute for mammals, thus achieving compression of information.\footnote{The concept of class-inclusion hierarchies with inheritance of attributes is quite fully developed in object-oriented programming, which originated with the Simula programming language \cite{birtwistle_etal_1973} and is now widely adopted in modern programming languages.}

This idea may be generalised to cross-classification, where any one entity or class may belong in one or more higher-level classes that do not have the relationship superclass/subclass, one with another. For example, a given person may belong in the class `woman' and `doctor' although `woman' is not a subclass of `doctor' and {\em vice versa}.

\subsubsection{Part-whole hierarchies}\label{part-whole_hierarchies_section}

Another widely-used idea is the {\em part-whole hierarchy} in which a given entity or class of entities is divided into parts and sub-parts through as many levels as is needed. Here, ICMUP may be achieved because two or more parts of a class such as `car' may share the overarching structure in which they all belong. So, for example, each wheel of a car, the doors of a car, the engine or a car, and so on, all belong in the same encompassing structure, `car', and it is not necessary to repeat that enveloping structure for each individual part.

\subsubsection{SP-multiple-alignment}\label{sp-multiple-alignment_section}

The seventh version of ICMUP, the {\em SP-multiple-alignment} construct outlined in \ref{sp-multiple-alignment_appendix}, encompasses all the preceding six versions of ICMUP and much more besides.

How the preceding six versions of ICMUP may be modelled within the SP-multiple-alignment framework is described in \cite[Appendix B]{sp_software_engineering}. The strengths and potential of the SP-multiple-alignment construct in modelling aspects of human intelligence and the representation of knowledge is outlined in \cite{sp_ghlai_2017} with pointers to where fuller information may be found. The potential of this construct in modelling aspects of mathematics is described in \cite[Chapter 10]{wolff_2006}.

\section{Mathematics as information compression via the matching and unification of patterns}\label{mathematics_as_icmup_section}

The first step in the argument for MICMUP, depends on evidence that mathematics is fundamentally about the compression of information. The following subsections present evidence in support of this idea.

In what follows, ICMUP may be seen to have an impact on the structuring of mathematics, and on the dynamics of mathematical calculation or inference.

\subsection{An example of information compression via mathematics}\label{example_of_ic_via_maths_section}

The equation $s = (gt^2) / 2$, one of several that can be derived from Newton's Second Law of Motion, is a very compact means of representing any table, including large ones, showing the distance travelled by a falling object ($s$) in a given time since it started to fall ($t$), as illustrated in Table \ref{distance_time_table}.\footnote{Of course, the law does not work for something like a feather falling in air.} The constant, $g$, is the acceleration due to gravity---about $9.8 m / s^2 $. That small equation would represent the values in the table even if it was a 1000 times or a million times bigger, and so on. Likewise for other equations such as $a^2 + b^2 = c^2$, $PV = k$, $F = q(E + v \times B)$, and so on.

\begin{table}[H]
\centering
\begin{tabular}{|r|r|} \hline
\em Distance (m)    & \em Time (sec) \\ \hline
\hline
0.0                 &      0 \\ \hline
4.9                 &      1 \\ \hline
19.6                &      2 \\ \hline
44.1                &      3 \\ \hline
78.5                &      4 \\ \hline
122.6               &      5 \\ \hline
176.5               &      6 \\ \hline
240.3               &      7 \\ \hline
313.8               &      8 \\ \hline
397.2               &      9 \\ \hline
490.3               &     10 \\ \hline
593.3               &     11 \\ \hline
706.1               &     12 \\ \hline
828.7               &     13 \\ \hline
961.1               &     14 \\ \hline
1103.2              &     15 \\ \hline
1255.3              &     16 \\ \hline
{\em Etc}           &    {\em Etc} \\ \hline
\end{tabular}
\caption{The distance travelled by a falling object (metres) in a given time since it started to fall (seconds).}
\label{distance_time_table}
\end{table}

To make these points, it is not strictly necessary to show Table \ref{distance_time_table}. But the table helps to emphasise the contrast between the potentially huge volumes of data in such a table and the small size of the equation which describes those data---and, correspondingly, the potentially high levels of information compression that may be achieved with ordinary mathematics which is not specialised for compression of information.

\subsection{How ICMUP may be seen in the structure and workings of mathematics}\label{maths_as_ic_techniques_section}

The subsections that follow describe how some of the basic principles and techniques for the compression of information that were outlined in Section \ref{seven_techniques_for_icmup_section} may be seen in the structure and workings of mathematics.

In themselves, these examples do not prove that mathematics may be understood as being entirely devoted to the compression of information. But there are reasons to think that compression of information is fundamental in mathematics:

\begin{itemize}

    \item Since the techniques to be described are low-level techniques that are part of the foundations of mathematics and widely used in more complex forms of mathematics, it seems likely that mathematics may indeed be understood in its entirety as ICMUP.

    \item As described in Section \ref{xor_examples_section}, the workings of simple logical functions, including the NAND logical function, may be understood in terms of ICMUP. Since it is widely accepted that, in principle, the computational heart of any general-purpose digital computer may be constructed entirely from NAND gates \cite{nisan_schocken_2005}, it appears that, within the bounds imposed by computational complexity, ICMUP has the generality to support any kind of computation, including mathematical computations.

\end{itemize}

\subsection{Basic ICMUP}\label{basic_icmup_examples_section}

The `basic' version of ICMUP, ``basic ICMUP''(Section \ref{basic_icmup_section}) may be seen in mathematics whenever one identifier is matched with another, with implicit unification of the two.

\subsubsection{The matching and unification of identifiers}\label{mu_identifiers_examples_section}

In mathematics, ICMUP may be seen wherever there is a need to invoke a named entity. If, for example, we want to calculate the value of $z$ from these equations: $x = 4$; $y = 5$; $z = x + y$, we need to match the identifier $x$ in the third equation with the identifier $x$ in the first equation, and to unify the two so that the correct value is used for the calculation of $z$. Likewise for $y$.

In a similar way if we wish to invoke or `call' a function such as `$\log x$' (the logarithm of a number), there must be a match between the name of the function in the call to the function (such as `$\log 1000$' and the name of the function in its definition, `$\log x$'. Unification of the call to the function with the definition of the function may be seen to have the effect of assigning the number in the call ($1000$ in this example) to the variable $x$ in the definition of the function.

\subsubsection{The execution of a function}\label{execution_of_function_examples_section}

At an abstract level, any function may be seen as a table in which each row shows the connection between one or more input values and one or more output values. And simple functions, such as a one-bit adder, may be specified in exactly that way, as shown in Table \ref{one-bit_adder_definition_table}.

\begin{table}[H]
\centering
\begin{tabular}{|r|r|r|r|} \hline
\em Input (1) & \em Input (2) & \em Sum & \em Carry \\ \hline
\hline
1 & 1 & 0 & 1 \\ \hline
1 & 0 & 1 & 0 \\ \hline
0 & 1 & 1 & 0 \\ \hline
0 & 0 & 0 & 0 \\ \hline
\end{tabular}
\caption{A table to define a function for the addition of two one-bit numbers in binary arithmetic, with provision for the carrying out of one bit.}
\label{one-bit_adder_definition_table}
\end{table}

In the workings of this adder, basic ICMUP may be seen, for example, in the matching and unification of input values `$1$' and `$0$' with corresponding values in `input' columns of the table. In this case, the matches which achieve the greatest compression (both `$1$' and `$0$' in one row) will be to select the second row in the table, with the sum `$1$' and the carry digit `$0$', which are of course the correct outputs for those two inputs.

\subsubsection{Matching and unification of patterns with Peano's axiom for natural numbers}\label{mup_peanos_nn_examples_section}

The sixth of Peano's axioms for natural numbers---for every natural number $n$, $S(n)$ is a natural number---provides the basis for a succession of numbers: $S(0)$, $S(S(0))$, $S(S(S(0)))$~..., itself equivalent to unary numbers in which $1 = /$, $2 = //$, $3 = ///$,~and so on. Here, $S$ at one level in the recursive definition is repeatedly matched and unified with $S$ at the next level.

\subsection{Chunking-with-codes}\label{chunking_codes_examples_section}

This subsection describes aspects mathematics that may be seen to exemplify the chunking-with-codes technique for information compression, as described in Section \ref{chunking-with-codes_section}.

\subsubsection{Named functions}\label{named_functions_examples_section}

If a body of mathematics is repeated in two or more parts of something larger then it is natural to declare it once as a named `function', where the body of the function may be seen as a `chunk' of information, and the name of the function is its `code' or identifier. This avoids the need to repeat that body of mathematics in two or more places.

An example of this kind of thing is the calculations needed to find the square root of a number, often provided as a ready-made square-root function with the non-alphabetic name `$\sqrt x$'. That name may be used to invoke the function wherever it is needed, like this: `$\sqrt{16}$'. Similar things may be done with functions such as `$\sin(x)$', `$\cos(x)$', and `$\log(x)$'.

Although they are not commonly seen as `functions', all of the operations of addition, subtraction, multiplication, the power notation, and division, may be cast in that mould as, for example, `{\em plus}({\em x,y})', `{\em subtract}({\em x,y})', and so on. As such, they may be seen as examples of chunking-with-codes and schema-plus-correction (Section \ref{schema_correction_examples_section}). As we shall see in Section \ref{run_length_coding_examples_section}, they may also be seen as examples of run-length coding.

\subsubsection{The number system}\label{number_system_example_section}

Number systems with bases greater than 1, like the binary, octal, decimal and hexadecimal number systems, may all be seen to illustrate the chunking-with-codes technique for compressing information. For example:

\begin{itemize}

    \item A unary number like `///////' may be referred to more briefly in the decimal system as `7'. Here, `///////' is the chunk and `7' is the code.

    \item \sloppy A unary number like `/////////////////' may be split into two parts: `//////////' and `///////'. Then, in the decimal system, the first part may be represented by `1' and the second part by `7', giving us the decimal number `17'. The convention is that the right-most digit represents numbers less than 10, and the next digit to the left represents the number of 10s.

    \item Of course, this `positional' system can be extended so that a digit in the third position from the right represents the number of 100s, a digit in the fourth position represents the number of 1000s, and so on.

\end{itemize}

Here, we can see how the chunking-with-codes technique allows us to eliminate the repetition or redundancy that exists in all unary numbers except `/'. This means that large numbers, like 2035723, may be expressed in a form that is very much more compact than the equivalent unary number.

\subsection{Schema-plus-correction}\label{schema_correction_examples_section}

Most functions in mathematics, like those mentioned above, are not only examples of chunking-with-codes: they are also examples of the schema-plus-correction device for compressing information. This is because they normally require input via one or more `arguments' or `parameters'. For example, the square root function needs a number like $49$ for it to work on. Without that number, the function is a very general `schema' for solving square root problems. With a number like $49$, which may be regarded as a `correction' to the schema, the function becomes focussed much more narrowly on finding the square root of $49$.

\subsection{Run-length coding}\label{run_length_coding_examples_section}

Run-length coding appears in various forms in mathematics, often combined with other things. The key idea is that some entity, pattern, or operation is repeated two or more times in an unbroken sequence. Here are some examples:

\begin{itemize}

    \item Since all numbers with bases above 1 may be seen to be compressed representations of unary numbers (Section \ref{number_system_example_section}), unary numbers may be regarded as more fundamental than non-unary numbers. If that is accepted, then, for example, `$3 + 7$' may be seen as a shorthand for the repeated process of transferring one unary digit from a group of seven unary digits to a group of three unary digits. Thus the expression `$+ 7$' within `$3 + 7$' may be seen as an example of run-length coding.

        Subtraction may be interpreted in a similar way when a smaller number is subtracted from a larger number.

    \item Multiplication is repeated addition. So, for example, `$3 \times 10$' is the 10-fold repetition of the operation `$x + 3$', where `$x$' starts with the value `$0$'. Then `$\times 10$' within `$3 \times 10$' may be seen as run-length coding. Since addition is itself a form of run-length coding (as described in the preceding bullet point), multiplication may be seen as run-length coding on two levels.

    \item Division of a larger number by a smaller one (eg, `$12 / 3$') is repeated subtraction which, as with multiplication, may be seen as run-length coding. Of course there will be a `remainder' if the larger number is not an exact multiple of the smaller number. As with addition as a part of multiplication, subtraction as a part of division means that division may be seen as run-length coding on two levels.

    \item The power notation (eg, `$10^9$') is repeated multiplication, and is thus another example of run-length coding. Since multiplication, as repeated addition, is a form of run-length coding, and since addition may be seen as run-length coding (the first bullet point above), the power notation may be seen as run-length coding on three levels!

    \item A factorial (eg, `$25!$') is repeated multiplication and subtraction.

    \item The bounded summation notation (eg, `$\sum_{i = 1}^{5}\frac{1}{i}$') and the bounded power notation (eg, `$\prod_{n=1}^{10}\frac{n}{n-1}$') are shorthands for repeated addition and repeated multiplication, respectively. In both cases, there is normally a change in the value of one or more variables on each iteration, so these notations may be seen as a combination of run-length coding and schema-plus-correction.

    \item In matrix multiplication, `$AB$', for example, is a shorthand for the repeated operation of multiplying each entry in matrix `$A$' with the corresponding entry in matrix `$B$'.

\end{itemize}

\subsection{Class-inclusion hierarchies}\label{class-inclusion_hierarchy_examples_section}

Classes and subclasses (Section \ref{class-inclusion_hierarchies_section}) feature in mathematics as `sets', both as a sometimes-disputed foundation for mathematics and as a branch of mathematics.

The notion of `inheritance' does not have the prominence in set theory that it does in object-oriented programming, but, nevertheless, ICMUP may be seen in other concepts associated with sets, described in Section \ref{logic_examples_section}.

\subsection{Part-whole hierarchies}\label{part-whole_hierarchy_examples_section}

It seems that part-whole hierarchies are not much used in mathematics, except perhaps in set theory, but, as we shall see in Section \ref{computing_examples_section}, they are quite prominent in the mathematics-related discipline of computing.

\subsection{SP-multiple-alignment}\label{sp_ma_examples_section}

Preliminary work described in \cite[Chapter 10]{wolff_2006} shows that the SP system, with SP-multiple-alignment centre-stage, has potential to model mathematical constructs and mathematical processes. This should not be altogether surprising since, as noted in Section \ref{sp-multiple-alignment_section}, SP-multiple-alignments can do everything that can be done with the six variants of ICMUP described in Sections \ref{basic_icmup_section} to \ref{part-whole_hierarchies_section}, and it provides for their seamless integration too.

Other reasons for believing that the SP system has potential to model many and perhaps all concepts and processes in mathematics are:

\begin{itemize}

    \item The generality of information compression as a means of representing knowledge in a succinct manner.

    \item The central role of information compression in the SP-multiple-alignment framework.

    \item The versatility of the SP-multiple-alignment framework in aspects of intelligence and the representation of knowledge (\ref{strengths_of_the_sp_system_appendix}).

    \item The close connection that is known to exist between information compression and concepts of inference and probability (Section \ref{ic_inference_probabilities_section}).

\end{itemize}

\subsection{Some equations}\label{some_equations_examples_section}

It seems that most equations that have become established in mathematics and science may be interpreted in terms of some combination of the techniques for compressing information described in Section \ref{seven_techniques_for_icmup_section}. Thus:

\begin{itemize}

    \item Einstein's equation, $E = mc^2$, illustrates run-length coding in its power notation ($c^2$) and in the multiplication of $m$ with $c^2$.

    \item Newton's equation, $s = (gt^2) / 2$, that featured in Section \ref{example_of_ic_via_maths_section}, illustrates run-length coding in its power notation ($t^2$), in the multiplication of $g$ with $t^2$, and in the division of $(gt^2)$ by $2$.

    \item Pythagoras's equation, $a^2 + b^2 = c^2$, illustrates run-length coding via the power notation in $a^2$, $b^2$, and $c^2$, and via the addition of $b^2$ to $a^2$ (the first bullet point in Section \ref{run_length_coding_examples_section}).

    \item Boyle's law, $PV = k$, illustrates run-length coding in the multiplication of $P$ by $V$.

    \item The charged particle equation, $F = q(E + v \times B)$, illustrates run-length coding in the multiplication of $v$ by $B$, in the multiplication of $(E + v \times B)$ by $q$, and in the addition of $v \times B$ to $E$.

    \item One of special relativity's equations for time dilation, $\Delta t^\prime = \Delta t / \sqrt {1 - v^2 / c^2}$, illustrates chunking-with-codes and schema-plus-correction in its use of the square root function, and it illustrates run-length coding in the division of $v^2$ by $c^2$, in the subtraction of $v^2 / c^2$ from $1$, and in the division of $\Delta t$ by $\sqrt {1 - v^2 / c^2}$.

    \item In its use of bounded summation ($\sum$), Shannon's equation for entropy, $H = - \sum_{i} p_i \log_2 (p_i)$, illustrates a combination of run-length coding and schema-plus-correction (as noted in Section \ref{run_length_coding_examples_section}). It also illustrates chunking-with-codes in its use of the $\log_2$ notation.

\end{itemize}

Since addition, subtraction, multiplication, the power notation, and division, may each be seen as an example of chunking-with-codes and schema-plus-correction (Sections \ref{chunking_codes_examples_section} and \ref{schema_correction_examples_section}), as well as run-length coding (Section \ref{run_length_coding_examples_section}), the same can be said about the appearance of those notations in each of the examples above.

\section{Mathematics-related disciplines as information compression via the matching and unification of patterns}\label{mathematics-related_disciplines_as_icmup_section}

It seems that, to a large extent, what has been said about mathematics in Section \ref{mathematics_as_icmup_section} also applies to the mathematically-related disciplines of logic and computing.\footnote{Where computing has its modern sense of computation by machine.} The following two subsections present some examples in support of that idea.

\subsection{Logic}\label{logic_examples_section}

Subsections that follow describe some evidence for ICMUP in logic.

\subsubsection{XOR and other logical operations}\label{xor_examples_section}

The XOR logical function, and other simple logical functions, may be defined and interpreted in much the same way as the one-bit adder shown in Table \ref{one-bit_adder_definition_table}, as shown in Table \ref{xor_definition_table}.

\begin{table}[H]
\centering
\begin{tabular}{|r|r|r|} \hline
\em Input (1) & \em Input (2) & \em Output \\ \hline
\hline
1 & 1 & 0 \\ \hline
0 & 1 & 1 \\ \hline
1 & 0 & 1 \\ \hline
0 & 0 & 0 \\ \hline
\end{tabular}
\caption{A table to define the XOR logical function.}
\label{xor_definition_table}
\end{table}

As with the one-bit adder, the operation of the XOR function may be understood in terms of basic ICMUP. Input values such as $1$ (first) and $0$ (second) may be matched and unified with values in the corresponding `input' columns of the table. With those two input values, the third row is selected because it yields most matches---which, with unification, also means the greatest compression of information. And of course the third row yields the correct output value, which in this example is $1$.

There are two points of interest here:

\begin{itemize}

    \item {\em The XOR Function and Artificial Neural Networks}. As is well known, Marvin Minsky and Seymour Papert \citeyearpar{minsky_papert_1969} demonstrated that basic perceptrons of the kind that were available in the late 1960s could not produce correct results with the XOR function, a demonstration which, for a time, led to a fall in interest in artificial neural networks.

    \item {\em The Generality of the NAND Logical Function}. As noted in Section \ref{maths_as_ic_techniques_section}, the fact that the NAND logical function may, like XOR and other simple logical functions, be understood in terms of ICMUP, and the generally-accepted idea that the computational heart of any general-purpose computer may, in principle, be constructed entirely from NAND gates, provide evidence in support of the idea that compression of information is fundamental in all kinds of computation including mathematical computations.

\end{itemize}

\subsubsection{Deriving a set from a multiset}\label{deriving_set_from_multiset_section}

In logic and mathematics, a `multiset' or `bag' is like a set but any element within the multiset may be repeated as, for example, in the multiset \{a, b, a, c, b, b, c, a, c\}.

Conversion of any such multiset into the corresponding set means matching each element within the multiset with every other element and, wherever a match is found, unifying the two elements, including elements that are the result of earlier unifications, thus achieving ICMUP. In this case, the multiset \{a, b, a, c, b, b, c, a, c\} is reduced to the set \{a, b, c\}.

\subsubsection{The union and intersection of sets}\label{union_and_intersection_of_sets_section}

In much the same way that a set may be derived from a multiset (Section \ref{deriving_set_from_multiset_section}), the union and intersection of two sets may be found by the matching and unification of elements, yielding a reduction in the overall size of the two sets when unification has been achieved. Thus, for example, the union of the sets \{b, f, d, a, c, e\} and \{e, g, i, f, d, h\} is \{a, b, c, d, e, f, g, h, i\}, with the intersection \{d, e, f\}. In accordance with ICMUP, the union is smaller than the two sets from which it was derived.

\subsubsection{ICMUP in Prolog}\label{icmup_in_prolog_section}

Further evidence for the significance of ICMUP in logic is that systems like Prolog---a computer-based version of logic---may be seen to function largely via the matching and merging of patterns.

Here, the meaning of `unification' in Prolog---comparing two terms to see if they can be made to represent the same structure---is quite close to the meaning of `unification' in this paper.

\subsubsection{Versatility in reasoning with the SP system}\label{versatility_in_reasoning_in_sp_system_section}

Since SP-multiple-alignment is a generalised form of ICMUP (Section \ref{sp-multiple-alignment_section}), and since SP-multiple-alignment is an important part of the SP system, it is pertinent to say that the SP Computer Model demonstrates several kinds of reasoning including: one-step `deductive' reasoning; chains of reasoning; abductive reasoning; reasoning with probabilistic networks and trees; reasoning with `rules'; nonmonotonic reasoning and reasoning with default values; Bayesian reasoning with `explaining away'; causal reasoning; reasoning that is not supported by evidence; the inheritance of attributes in class hierarchies; and inheritance of contexts in part-whole hierarchies (\cite[Section 10]{sp_extended_overview}, \cite[Chapter 7]{wolff_2006}).

Because of the probabilistic nature of the SP system, these forms of reasoning are probabilistic, although some of them, such as one-step `deductive' reasoning, have the all-or-nothing character of traditional forms of logic. Nevetheless, if it is accepted that logic, like mathematics, is probabilistic at a deep level---for reasons given in Section \ref{ic_inference_probabilities_section}---then the above-mentioned strengths of the SP system in probabilistic reasoning may be seen as further evidence for the importance of ICMUP in logic.

\subsection{Computing}\label{computing_examples_section}

As with logic, it seems likely that, since computing is closely related to mathematics, it may, like mathematics, be understood in terms of ICMUP. Evidence in support of that view is presented in subsections that follow.

\subsubsection{Matching and unification of patterns in definitions of `computing'}

Emil Post's \citeyearpar{post_1943} ``Canonical System'', which is recognised as a definition of `computing' that is equivalent to a universal Turing machine, may be seen to work largely via the matching and unification of patterns \cite[Chapter 4]{wolff_2006}.

Much the same is true of the workings of the transition function in a universal Turing machine. This is essentially a look-up table like that shown in Table \ref{look-up_table}.

\begin{table}[H]
\centering
\begin{tabular}{|r|r|r|r|} \hline
\em Input (1) & \em Input (2) & \em Output (1) & \em Output (2) \\ \hline
\hline
$s_0$  &  1 & $s_0$ & {\Large \guillemotright} \\ \hline
$s_0$  &  0 & $s_1$ & 1 \\ \hline
$s_1$  &  1 & $s_1$ & {\Large \guillemotleft} \\ \hline
$s_1$  &  0 & $s_2$ & {\Large \guillemotright} \\ \hline
\end{tabular}
\caption{An example of a transition function in a universal Turing machine, represented as a look-up table, as described in the text. {\em Key:} `{\Large \guillemotright}' means ``move the read/write head one place to the right''; `{\Large \guillemotleft}' means ``move the read/write head one place to the left''. Based on the example in \cite[Section 2]{barker-plummer_2012}, with permission. }
\label{look-up_table}
\end{table}

Much as with the examples described in Sections \ref{execution_of_function_examples_section} and \ref{xor_examples_section}, ICMUP may be seen, for example, in the matching and unification of input values `$s_1$' and `1' with corresponding values in the input columns of the table. In this case, the effect will be to select the third row in the table, with the output values `$s_1$' and `{\Large \guillemotleft}'---which mean ``Set the state of the machine to `$s_1$' and move the read/write head of the machine one place to the left''.

In a similar way, ICMUP may be seen in the workings of the NAND logical function which, as noted in Sections \ref{maths_as_ic_techniques_section} and \ref{xor_examples_section} may in principle provide the computational heart of any general-purpose digital computer.

\subsubsection{Some other examples of ICMUP in computing}

Here, in brief, are some other putative examples of ICMUP in computing:

\begin{itemize}

    \item {\em Basic ICMUP}. As in mathematics (Section \ref{basic_icmup_examples_section}), basic ICMUP may be seen in computing in the matching of identifiers for variables and in calls to functions.

    \item {\em Chunking-With-Codes and Schema-Plus-Correction}. Again, as in Section \ref{chunking_codes_examples_section}, named functions in computing may be seen as examples of the chunking-with-codes version of ICMUP, and as in Section \ref{schema_correction_examples_section}, functions with parameters may be seen as examples of the schema-plus-correction version of ICMUP.

    \item {\em Run-Length Coding}. As in mathematics (Section \ref{run_length_coding_examples_section}), run-length coding may be seen in computing in the basic arithmetic functions. It may also be seen in iteration statements like {\em while ...}, {\em do ...~while ...}, {\em for ...}, or {\em repeat ...~until ...}. And it may be seen in the use of recursion in functions such as {\em factorial(x)} for the calculations of the factorial of any number.

    \item {\em Class-Inclusion and Part-Whole Hierarchies}. In computing, the creation of classes and hierarchies of classes is supported in such object-oriented programming languages as Simula, Smalltalk, C++, and many more. Part-while hierarchies are also prominent in software. In both cases, ICMUP has a role to play, much as described in Sections \ref{class-inclusion_hierarchies_section} and \ref{part-whole_hierarchies_section}.

    \item {\em Retrieving Data From Computer Memory}. It is true that electronic circuits provide the mechanism for finding an address in computer memory but, at a more abstract level, the process may be seen as one of searching for a match between the address held in the CPU and the corresponding address in computer memory. When a match has been found between the address in the CPU and the corresponding address in memory, there is implicit unification of the two.

    \item {\em Query-by-Example}. A popular technique for retrieving information from databases, query-by-example, is essentially a process of finding good matches between a query pattern and patterns in the database, with unification of the best matches.

\end{itemize}

\section{Information compression, inference, and probabilities}\label{ic_inference_probabilities_section}

The main focus of this paper is on MICMUP, but it is relevant to mention that it has been recognised for some time that there is an intimate connection between information compression and concepts of inference and probability, as described in \cite{shannon_weaver_1949}, in Ray Solomonoff's {\em Algorithmic Probability Theory} (APT) \cite{solomonoff_1964,solomonoff_1997}, and in the closely-related AIT \cite{li_vitanyi_2014}. Information compression and concepts of inference and probability may be seen as two sides of the same coin.

The close connection between those things makes sense in terms of ICMUP (Section \ref{seven_techniques_for_icmup_section}):

\begin{itemize}

    \item A pattern that repeats is one that, via inductive reasoning, we naturally regard as a guide to what may happen in the future.

    \item A pattern that repeats is one that, via the merging or unification of patterns, may yield compression of information.

    \item A partial match between one pattern and another can be the basis for infering the occurrence of the unmatched parts, a form of inference that is sometimes called {\em prediction by partial matching} \cite{teahan_alhawiti_2015}.

\end{itemize}

What has this got to do with mathematics? It would take us too far afield to discuss this issue in any depth. A few brief remarks are made here. The close connection between information compression and concepts of inference and probability, and evidence for MICMUP presented in this paper, suggests that:

\begin{itemize}

    \item Notwithstanding the apparent certainty of equations like $2 + 2 = 4$, mathematics may be seen to be fundamentally probabilistic.

    \item In view of the important role that mathematics has in the making of inferences in science and elsewhere, and notwithstanding the apparent certainties of many of those inferences, MICMUP may be seen as a driver for the making of `exact' inferences.

\end{itemize}

Regarding the first point, a probabilistic foundation for mathematics is consistent with the discovery of randomness in number theory:

\begin{quote}

    ``I have recently been able to take a further step along the path laid out by G{\"o}del and Turing. By translating a particular computer program into an algebraic equation of a type that was familiar even to the ancient Greeks, I have shown that there is randomness in the branch of pure mathematics known as number theory. My work indicates that---to borrow Einstein's metaphor---God sometimes plays dice with whole numbers.'' \cite[p.~80]{chaitin_1988}.

\end{quote}

As indicated in this quotation, randomness in number theory is closely related to G{\"o}del's incompleteness theorems. These are themselves closely related to the phenomenon of recursion, a feature of many formal systems, several of Escher's pictures, and much of Bach's music, as described in some detail by Douglas Hofstadter in {\em G{\"o}del, Escher, Bach: An Eternal Golden Braid} \cite{hofstadter_1980}.

Again with regard to the first point, the SP system, which is dedicated to information compression, has clear strengths in the making of uncertain inferences and the calculation of associated probabilities (\cite[Section 4.4]{sp_extended_overview} and \cite[Sections 3.7 and Chapter 7]{wolff_2006}). It seems possible that, with further development, the SP system may have strengths to rival conventional statistics.

With regard to the second point, it seems possible that, although mathematics may be fundamentally probabilistic, it may, with appropriate data or under appropriate conditions, deliver results where the associated probabilities are at or very close to 0 or 1. This kind of possibility is discussed briefly in \cite[Section 6.3]{sp_extended_overview} and in \cite[Sections 10.4.5 and 10.4.6]{wolff_2006}.

\section{Information compression, mathematics, and human learning, perception, and cognition}\label{ic_maths_hlpc_section}

As indicated in the Introduction, and in Sections \ref{the_big_picture_section} and \ref{novelty_of_maths_as_ic_section}, MICMUP makes sense:

\begin{itemize}

    \item If mathematics is viewed as a form of human thought, by contrast, for example, with the Platonist view that mathematical entities:

    \begin{quote}

        ``...~are not merely formal or quantitative structures imposed by the human mind on natural phenomena, nor are they only mechanically present in phenomena as a brute fact of their concrete being. Rather, they are numinous and transcendent entities, existing independently of both the phenomena they order and the human mind that perceives them.'' \cite[pp.~95--96]{hersh_1997}.

    \end{quote}

    \item And if human thought, like other aspects of HLPC, is seen to be driven largely by processes of information compression, in accordance with evidence presented in \cite{sp_compression}, and in accordance with the principles at the heart of the SP Theory of Intelligence (Section \ref{sp_theory_section} and \ref{sp_system_appendix}).

\end{itemize}

\section{{S}o what?}\label{so_what_section}

While it may be accepted that mathematics may be understood as information compression via the matching and unification of patterns, readers may wonder what benefits or applications there may be, if any, for MICMUP and related ideas. Here are some possibilities:

\begin{itemize}

    \item {The Big Picture}. The evidence and arguments in this paper provide support for the Big Picture and its six components, outlined in Section \ref{the_big_picture_section}. In keeping with Ockham's Razor, the Big Picture is important in showing the potential of information compression as a unifying principle across a wide canvass.

    \item {The development of mathematics}. Since the MICMUP ideas in this paper have grown out of the SP Theory of Intelligence (Section \ref{introduction_section}), there is potential for augmenting mathematics with concepts and mechanisms from the SP System, especially SP-multiple-alignment and its associated mechanisms, and unsupervised learning via the building of grammars. Those concepts, together with the MICMUP concepts, may lead to such things as: radically new ways of creating mathematical conjectures or hypotheses (more below); a radically new approach to the proof of theorems, propositions, lemmas, and so on, via compression of information.

    \item {A new mathematics for science}. There is potential for the development of a new mathematics for science (NMFS) as outlined in Appendix \ref{new_mathematics_for_science_appendix}. Possibilities here include:

        \begin{itemize}

            \item Extending the range of things that mathematics may do in science so that it becomes a {\em universal framework for the representation and processing of diverse kinds of knowledge} (UFK) \cite[Section III]{sp_big_data} that may take over from diagrams, videos, and descriptions in natural language.

            \item Using mathematics as means of quantifying the {\em Simplicity} of any scientific theory, and its descriptive or explanatory {\em Power}, and thus facilitating quantitative comparisons amongst rival scientific theories.

            \item The automatic or semi-automatic creation of scientific theories from data \cite[Section 6.10.7]{sp_benefits_apps}.

            \item By providing a UFK for the description and processing of related but incompatible theories such as quantum mechanics and relativity, a NMFS has the potential to help iron out inconsistencies and facilitate the integration of such theories.

        \end{itemize}

    \item {Sources of hypotheses}. In view of the evidence presented in this paper and in \cite{sp_compression}, and evidence in support of the SP Theory of Intelligence (\ref{sp_system_appendix}), information compression, and more specifically ICMUP, SP-multiple-alignment and the SP Theory of Intelligence, are likely to be fertile sources of hypotheses in the study of: the foundations of mathematics, logic and computer science; concepts of inference and probability; science and scientific methods; human learning, perception, and cognition; and neuroscience.

\end{itemize}

In view of the many potential benefits and applications of the SP System (\ref{potential_benefits_and_appications_appendix}), there are reasons to anticipate more such benefits and applications from MICMUP concepts and associated ideas.

\section{Conclusion}

This paper describes a novel perspective on the foundations of mathematics: how mathematics may be seen to be largely about `information compression via the matching and unification of patterns' (ICMUP). `Mathematics as ICMUP' may be shortened to `MICMUP'.

ICMUP is itself a novel approach to information compression, couched in terms of non-mathematical primitives, as is necessary in any investigation of the foundations of mathematics.

This new perspective on the foundations of mathematics has grown out of an extensive programme of research developing the SP Theory of Intelligence and its realisation in the SP Computer Model, a system in which a generalised version of ICMUP---the powerful concept of SP-multiple-alignment---plays a central role.

These ideas may be seen to be part of a `Big Picture' comprising six areas of interest, with information compression as a unifying theme.

Seven variants of ICMUP are described in Section \ref{seven_techniques_for_icmup_section}.

In arguing for MICMUP, Section \ref{mathematics_as_icmup_section} shows first how mathematics may achieve compression of information. Then it shows how variants of ICMUP may be seen in widely-used structures and operations in mathematics.

Section \ref{mathematics-related_disciplines_as_icmup_section}, argues that, in a similar way, mathematics-related disciplines of logic and computing may be seen as ICMUP.

Section \ref{ic_inference_probabilities_section} discusses how the intimate relation between information compression and concepts of inference and probability may be seen as a driver for the making of mathematical inferences, how it relates to the already-established view that, at a fundamental level, mathematics is intrinsically probabilistic, and how that latter view may be reconciled with the all-or-nothing, `exact', forms of calculation or inference that are familiar in mathematics.

In Section \ref{ic_maths_hlpc_section}, it is argued that MICMUP makes sense if mathematics is seen as a form of human thought and if human thought, like other aspects of HLPC, is seen to be driven largely by processes of information compression, in accordance with much empirical evidence \cite{sp_compression}, and the principles at the heart of the SP Theory of Intelligence.

Section \ref{so_what_section} outlines many potential benefits and applications from MICMUP concepts and associated ideas.

\section*{Acknowledgements}

I'm grateful to Roger Penrose, John Barrow, and anonymous reviewers for very helpful comments on earlier versions of this paper. For helpful comments on drafts of a related paper, I'm grateful to Robert Thomas, Michele Friend, and Alex Paseau. I'm also grateful for discussion from time to time of some of the ideas in this paper with Tim Porter and Chris Wensley.

\section*{Appendices}

\appendix

\section{The SP Theory of Intelligence and the SP Computer Model}\label{sp_system_appendix}

As noted in the Introduction, much of the thinking in this paper derives from the {\em SP Theory of Intelligence} and its realisation in the {\em SP Computer Model}. And, as previously noted, the SP Theory aims to simplify and integrate observations and concepts across artificial intelligence, mainstream computing, mathematics, and HLPC.

The SP system is introduced in \cite{sp_intro_2018}, described at more length in \cite{sp_extended_overview}, and described much more fully in \cite{wolff_2006}. Several other papers from the SP programme of research, most with download links, may be found via
\href{http://www.cognitionresearch.org/sp.htm}{www.cognitionresearch.org/sp.htm}.

\begin{figure}[!htbp]
\centering
\includegraphics[width=0.9\textwidth]{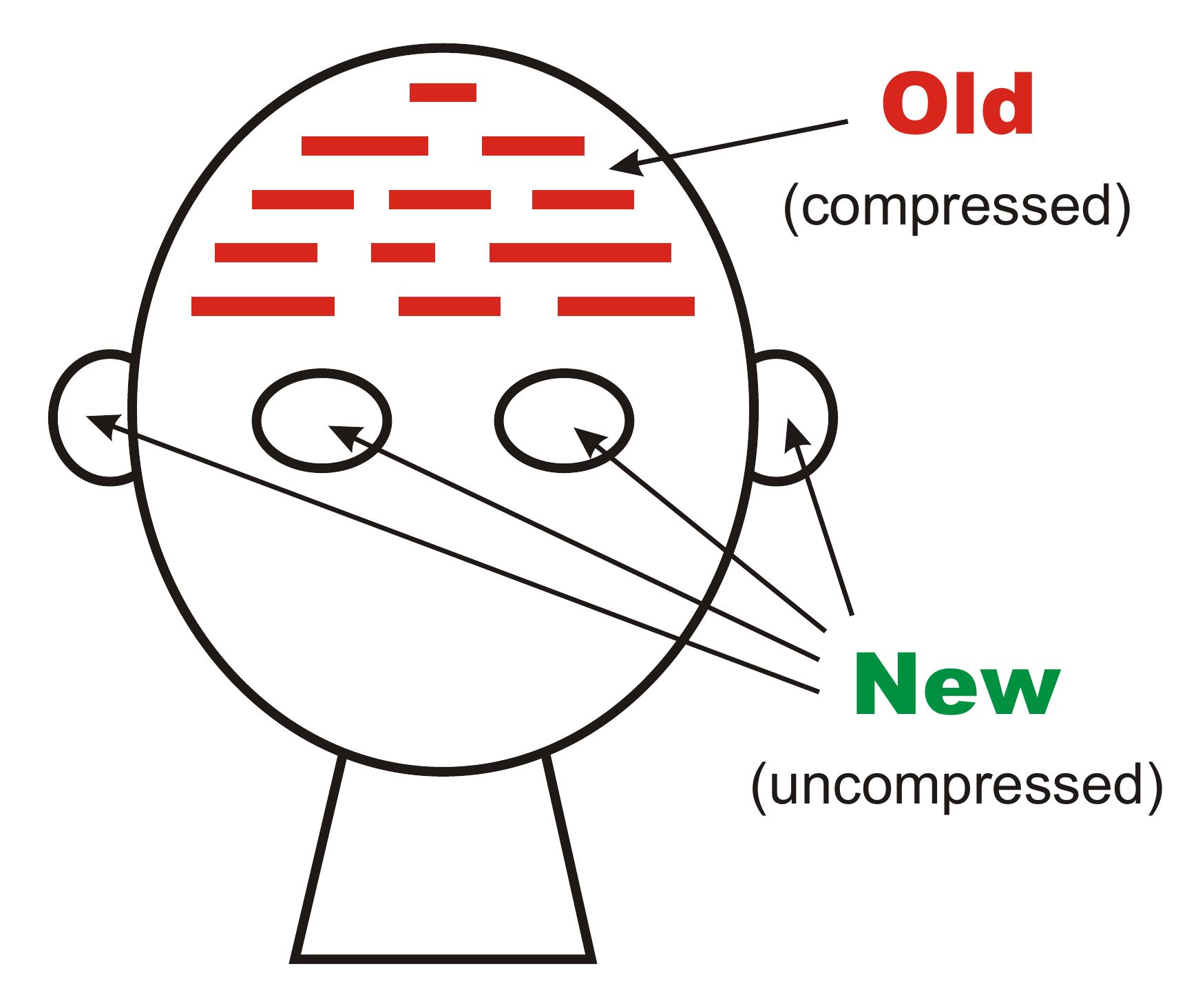}
\caption{Schematic representation of the SP system from an `input' perspective. Reproduced, with permission, from Figure 1 in \cite{sp_extended_overview}.}
\label{sp_input_perspective_figure}
\end{figure}

As shown schematically in Figure \ref{sp_input_perspective_figure}, the SP Theory is conceived as a brain-like system that receives {\em New} information via its senses and stores some or all of it, in compressed form, as {\em Old} information.

In the SP system, all kinds of knowledge are stored as arrays of atomic {\em SP-symbols} called {\em SP-patterns}. At present, the SP Computer Model works only with one-dimensional SP-patterns but it is envisaged that it will be generalised to work with two-dimensional SP-patterns, in addition to 1D SP-patterns.

\subsection{SP-multiple-alignment}\label{sp-multiple-alignment_appendix}

A key idea in the SP system is the concept of {\em SP-multiple-alignment} borrowed and adapted from the concept of `multiple alignment' in bioinformatics.

An example of a multiple alignment from bioinformatics is shown in Figure \ref{ma_dna_figure}. Here, five DNA sequences have been arranged in rows and, by judicious ``stretching'' of sequences in a computer, matching symbols have brought into line. A `good' multiple alignment is one with a relatively large number of matching symbols.

\begin{figure}[H]
\fontsize{10.00pt}{12.00pt}
\centering
{\bf
\begin{BVerbatim}
  G G A     G     C A G G G A G G A     T G     G   G G A
  | | |     |     | | | | | | | | |     | |     |   | | |
  G G | G   G C C C A G G G A G G A     | G G C G   G G A
  | | |     | | | | | | | | | | | |     | |     |   | | |
A | G A C T G C C C A G G G | G G | G C T G G A | G A
  | | |           | | | | | | | | |   |   |     |   | | |
  G G A A         | A G G G A G G A   | A G     G   G G A
  | |   |         | | | | | | | |     |   |     |   | | |
  G G C A         C A G G G A G G     C   G     G   G G A
\end{BVerbatim}
}
\caption{A `good' multiple alignment amongst five DNA sequences. Reproduced with permission from Figure 3.1 in \cite{wolff_2006}.}
\label{ma_dna_figure}
\end{figure}

The key difference between the concept of multiple alignment in bioinformatics and the concept of SP-multiple-alignment is that, in the latter, a `good' SP-multiple-alignment is one that allows one New SP-pattern (sometimes more than one) to be encoded economically in terms of one or more Old SP-patterns.

An example of an SP-multiple-alignment is shown in Figure \ref{parsing_kittens_figure}. Here, the New SP-pattern is the sentence `\texttt{ t w o k i t t e n s p l a y}' shown in row 0. Each of rows 1 to 8 shows an Old SP-pattern representing a grammatical structure, which in rows 1, 3 and 5 is a word. The overall effect of the SP-multiple-alignment is to analyse or parse the sentence into its constituent parts, each one marked with its grammatical category.

\begin{figure}[H]
\fontsize{07.00pt}{08.40pt}
\centering
{\bf
\begin{BVerbatim}
0                      t w o              k i t t e n     s                  p l a y           0
                       | | |              | | | | | |     |                  | | | |
1                      | | |         Nr 5 k i t t e n #Nr |                  | | | |           1
                       | | |         |                 |  |                  | | | |
2                      | | |    N Np Nr               #Nr s #N               | | | |           2
                       | | |    | |                         |                | | | |
3               D Dp 4 t w o #D | |                         |                | | | |           3
                |            |  | |                         |                | | | |
4            NP D            #D N |                         #N #NP           | | | |           4
             |                    |                             |            | | | |
5            |                    |                             |       Vr 1 p l a y #Vr       5
             |                    |                             |       |             |
6            |                    |                             |  V Vp Vr           #Vr #V    6
             |                    |                             |  | |                   |
7 S Num    ; NP                   |                            #NP V |                   #V #S 7
     |     |                      |                                  |
8   Num PL ;                      Np                                 Vp                        8
\end{BVerbatim}
}
\caption{The best SP-multiple alignment created by the SP Computer Model with a store of Old SP-patterns like those in rows 1 to 8 (representing grammatical structures, including words) and a New SP-pattern, `\texttt{ t w o k i t t e n s p l a y}', shown in row 0 (representing a sentence to be parsed). Adapted from Figures 1 in \protect\cite{wolff_sp_intelligent_database}, with permission.}
\label{parsing_kittens_figure}
\end{figure}

It turns out that the concept of SP-multiple-alignment within the SP system can do much more than the parsing of sentences, as indicated in the next subsection.

A New SP-pattern like `\texttt{t w o k i t t e n s p l a y}' in row 0 of Figure \ref{parsing_kittens_figure} would normally be a pattern that is derived relatively directly from the environment of a person or artificial system. Old patterns in the system would have been developed from previously-seen New patterns via unsupervised learning.

\subsection{Strengths and potential of the SP system}\label{strengths_of_the_sp_system_appendix}

Distinctive features and advantages of the SP system compared with other AI-related systems are described in \cite{sp_alternatives}.

A key strength of the SP system is the way in which the simple but powerful concept of SP-multiple-alignment may yield many different capabilities:

\begin{itemize}

    \item {\em Versatility in aspects of intelligence} including: unsupervised learning; the analysis and production of natural language; pattern recognition that is robust in the face of errors; pattern recognition at multiple levels of abstraction; computer vision; best-match and semantic kinds of information retrieval; planning; and problem solving), and also {\em versatility in several kinds of reasoning} including: one-step `deductive' reasoning; chains of reasoning;  abductive reasoning; reasoning with probabilistic networks and trees;  reasoning with ‘rules’;  nonmonotonic reasoning and reasoning with default values; Bayesian reasoning with `explaining away'; causal reasoning; reasoning that is not supported by evidence; the inheritance of attributes in class hierarchies; and inheritance of contexts in part-whole hierarchies;

    \item {\em Versatility in the representation of diverse kinds of knowledge} including: the syntax of natural languages; class-inclusion hierarchies (with or without cross classification); part-whole hierarchies; discrimination networks and trees; if-then rules; entity-relationship structures; relational tuples; and concepts in mathematics, logic, and computing, such as `function', `variable', `value', `set', and `type definition. The addition of two-dimensional SP-patterns to the SP Computer Model is likely to expand the representational repertoire of the SP system to structures in two-dimensions and three-dimensions, and the representation of procedural knowledge with parallel processing.

    \item {\em Seamless integration of diverse aspects of intelligence and diverse kinds of knowledge, in any combination}.

\end{itemize}

\begin{figure}[!hbt]
\centering
\includegraphics[width=0.9\textwidth]{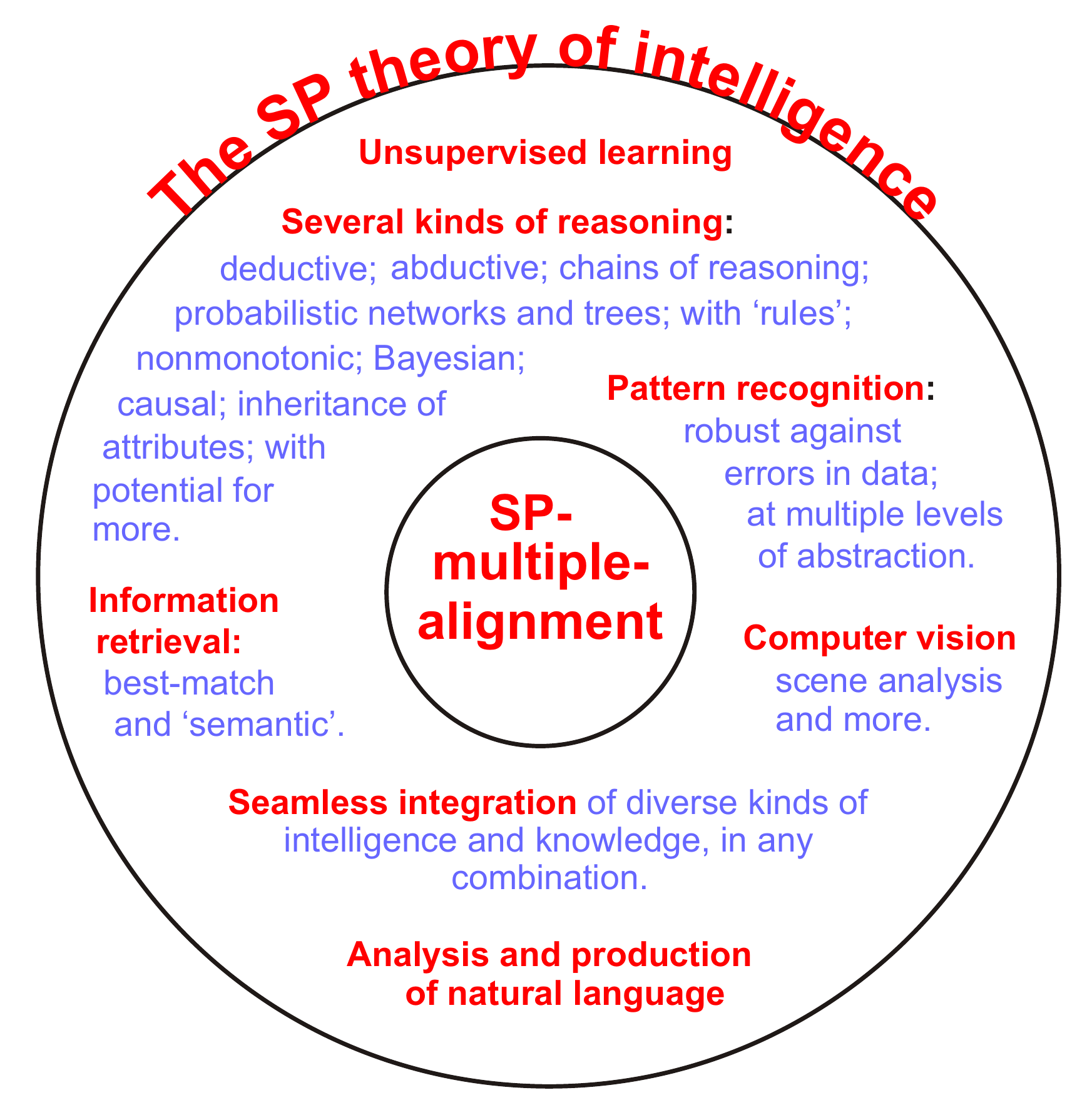}
\caption{A schematic representation of versatility and integration in the SP system, with SP-multiple-alignment centre stage.}
\label{versatility_integration_figure}
\end{figure}

Figure \ref{versatility_integration_figure} shows schematically how the SP system, with SP-multiple-alignment centre stage, exhibits versatility in its capabilities and their integration.

More detail may be found in \cite[Sections 4, 5, and 6]{sp_intro_2018} and in other sources cited there.

\subsection{The SP Machine}

\begin{figure}[!htbp]
\centering
\includegraphics[width=0.9\textwidth]{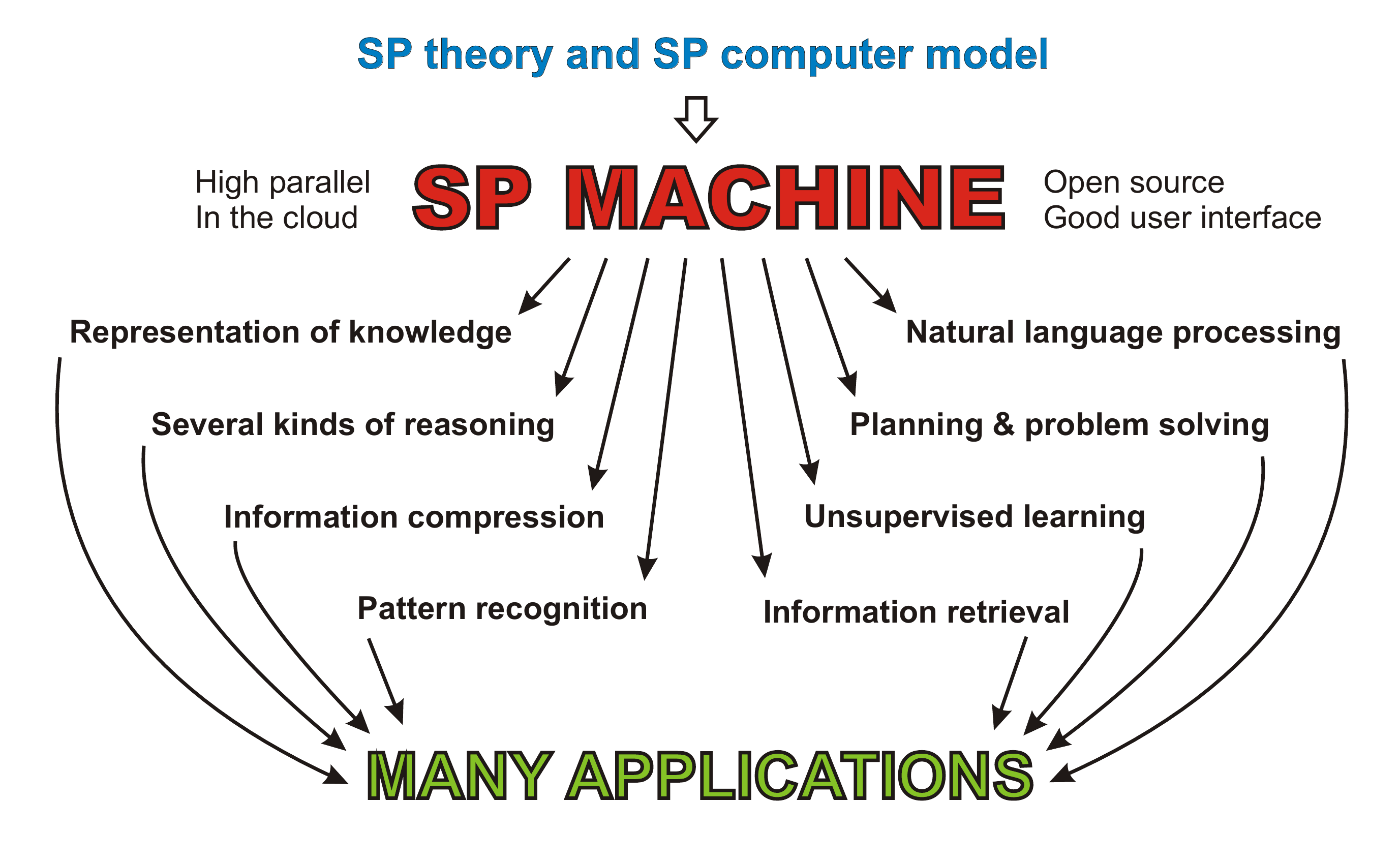}
\caption{Schematic representation of the development and application of the SP machine. Reproduced from Figure 2 in \cite{sp_extended_overview}, with permission.}
\label{sp_machine_figure}
\end{figure}

It is envisaged that the SP Computer Model will provide the basis for the development of a highly-parallel {\em SP Machine}, as shown schematically in Figure \ref{sp_machine_figure}. This projected development, described in \cite{palade_wolff_2018}, would be a vehicle for further research, and ultimately the basis for a system with the scale and robustness needed for industrial, commercial, and administrative applications.

\subsection{Potential benefits and applications}\label{potential_benefits_and_appications_appendix}

The SP system has potential in several areas of application including: helping to solve nine problems with big data; helping in the development of human-like intelligence in autonomous robots; helping in the understanding of human vision and in the development of computer vision; helping with medical diagnosis; functioning as a database system with intelligence;  and more.

There is more detail in \cite[Section 7]{sp_intro_2018} with pointers to where fuller information may be found.

\subsection{SP-Neural}\label{sp-neural_appendix}

Key concepts in the SP theory may be mapped on to structures of neurons and their interconnections in a version of the SP theory called {\em SP-Neural} \cite{spneural_2016}.

\section{The potential development of a {\em New Mathematics for Science}}\label{new_mathematics_for_science_appendix}

In Section \ref{so_what_section}, the potential is noted for developing some kind of {\em New Mathematics for Science} (NMFS), taking advantage of insights described in this paper into the nature of mathematics and its applications in science. Potential features of such an NMFS, with corresponding benefits, may include:

\begin{itemize}

    \item Rationalisation of mathematical notations, perhaps along the lines of a {\em universal framework for the representation and processing of diverse kinds of knowledge} (UFK), as described in \cite[Section III]{sp_big_data}. There seems to be potential to make mathematics more transparent in the representation of fundamentals such as ICMUP and its workings. There is a corresponding potential for mathematics to be easier to learn, to understand, and to use.

    \item Revision of such concepts as `proof' and `theorems', perhaps incorporating measures of information compression. Potential benefits here include a softening of the boundary between `exact', all-or-nothing styles of reasoning, and probabilistic kinds of reasoning, both of which are important in science.

    \item Development of procedures for the automatic or semi-automatic discovery of new results in mathematics.

    \item Incorporation of current and future versions of the SP system within the NMFS, with its strengths and potential in diverse aspects of intelligence (\ref{sp_system_appendix}, \cite[Sections 4, 5, and 6]{sp_intro_2018}). Such a development:

        \begin{itemize}

            \item May facilitate the automatic or semi-automatic development of new theories in science.

            \item May facilitate the quantitative evaluation of scientific theories, both existing theories and new ones.

            \item May broaden the scope of mathematics as a means of describing scientific observations and concepts succinctly,

            \item And it may provide a means of drawing inferences about scientific observations and concepts.

        \end{itemize}

\end{itemize}

Preliminary and tentative description of how an NMFS may be developed and what it may do is described in \cite{sp_maths_science}.

\section{Related issues}\label{related_issues_appendix}

This appendix considers briefly some issues related to the Big Picture (Section \ref{the_big_picture_section}).

\subsection{The apparent paradox of `decompression by compression'}\label{creading_redundancy_by_compression_appendix}

The idea that mathematics and related disciplines are largely, perhaps entirely, about compression of information seems to conflict with the undoubted fact that, with some simple mathematics or a simple computer program, it is possible to create data containing large amounts of repetition or redundancy.

This apparent inconsistency may be resolved via the concept of {\em decompression by compression}, described in \cite[Appendix B.1]{sp_compression}.

\subsection{Redundancy is often useful in the storage and processing of information}\label{redundancy_often_useful_appendix}

There is no doubt that informational redundancy---repetition of information---is often useful. For example, it is standard practice in computing to maintain two or more copies of important data, and redundancy in messages can provide a useful means of correcting errors. These kinds of uses of redundancy may seem to conflict with the idea that information compression---which means reducing redundancy---is fundamental in mathematics and related disciplines.

This issue and how it may be resolved is discussed in \cite[Appendix B.2]{sp_compression}.

\bibliographystyle{elsarticle-num}

\end{document}